\documentclass[final,12pt]{elsarticle}

\usepackage{graphicx}
\usepackage{amssymb}
\usepackage{amsmath}

\usepackage{lineno}

\usepackage{subcaption}
\usepackage{multirow}
\usepackage{CJK}
\usepackage[noend]{algpseudocode}
\usepackage{algorithm}
\usepackage{url}
\usepackage{color}
\usepackage{arydshln}

\usepackage{tikz}
\usetikzlibrary{patterns}
\usetikzlibrary{bayesnet}

\newcommand{\leaemph}[1]{\textsl{#1}}
\newcommand{\category}[1]{\textsc{#1}}    
\newcommand{\concept}[1]{\emph{#1}}       
\newcommand{\feature}[1]{\texttt{#1}}     

\definecolor{leaRed}{rgb}{0.722, 0, 0.278}

\usepackage{etoolbox}
\patchcmd{\pprintMaketitle}
  {\fi\hrule}
  {\fi\ifvoid\extrainfobox\else\unvbox\extrainfobox\par\vskip10pt\fi\hrule}
  {}{}
  
\newenvironment{extrainfo}
  {\global\setbox\extrainfobox=\vbox\bgroup\parindent=0pt }
  {\egroup}
\newsavebox\extrainfobox

\begin{document}
\begin{frontmatter}
\title{Categorization in the Wild: Generalizing Cognitive Models to Naturalistic 
Data  across Languages}


\author[a]{Lea Frermann \corref{cor1}}
\author[b]{Mirella Lapata}
\cortext[cor1]{Corresponding author. Email: uedi@frermann.de }

\address[a]{Amazon CoreAI}
\address[b]{University of Edinburgh}

\begin{abstract}
  Categories such as \category{animal} or \category{furniture} are
  acquired at an early age and play an important role in processing,
  organizing, and communicating world knowledge. Categories exist
  across cultures: they allow to efficiently represent the complexity
  of the world, and members of a community strongly agree on their
  nature, revealing a shared mental representation. Models of category
  learning and representation, however, are typically tested on data
  from small-scale experiments involving small sets of concepts with
  artificially restricted features; and experiments predominantly
  involve participants of selected cultural and socio-economical groups 
  (very often involving western native speakers of English such as~U.S. college
  students) . This work investigates whether models of categorization
  generalize (a)~to rich and noisy data approximating the environment
  humans live in; and (b)~across languages and cultures. We
  present a Bayesian cognitive model designed to jointly learn
  categories and their structured representation from natural language
  text which allows us to (a) evaluate performance on a large scale,
  and (b) apply our model to a diverse set of languages. We show that
  meaningful categories comprising hundreds of concepts and richly
  structured featural representations emerge across languages. Our
  work illustrates the potential of recent advances in computational
  modeling and large scale naturalistic datasets for cognitive science
  research.\end{abstract}

\begin{keyword}
\vspace{-0.5ex}
Categorization \sep Bayesian Modeling \sep Cognitive Modeling \sep Natural 
Language Processing
\end{keyword}
\begin{extrainfo}
\vspace{-1.5ex} {\it Conflict of interest}: None.\\
\end{extrainfo}

\end{frontmatter}


\section{Introduction}
\label{sec:intro}

Categories such as \category{animal} or \category{furniture} are
fundamental cognitive building blocks allowing humans to efficiently
represent and communicate the complex world around them. Concepts
(e.g.,~\concept{dog}, \concept{table}) are grouped into categories
based on shared properties pertaining, for example, to
their~\feature{behavior}, \feature{appearance}, or
\feature{function}.\footnote{We denote (superordinate level)
  \category{categories} (such as \category{animal} or
  \category{vehicle}) in small caps; (basic level) \concept{concepts}
  (such as \concept{cat} or \concept{car}) in italics, and
  \feature{feature types} (such as \feature{function} or
  \feature{behavior}) in \feature{true type}. Individual features
  (such as \{eats, sits, barks\}) are represented as \{lists\}.}
Categorization underlies other cognitive functions such as
perception~\cite{Schyns:1999,Goldstone:2003} or
language~\cite{waxman:properties,borovsky:learning}, and there is
evidence that categories are not only shaped by the world they
represent, but also by the language through which they are
communicated~\citep{Gopnik:1987,Waxman:1995}. Although mental
categories exist across communities and cultures, their exact
manifestations
differ~\citep{Malt:1995,Ji:2001,Unsworth:2005,Medin:2007}. For
example, American English speakers prefer taxonomic categorizations
(e.g., \concept{mouse},\concept{squirrel}) while Chinese speakers
tend to prefer to categorize objects relationally (e.g.,~\concept{tree},
\concept{squirrel}; \cite{Ji:2001}).

Given their prevalent function in human cognition, the acquisition and
representation of categories has attracted considerable attention in
cognitive science, and numerous theories have
emerged~\citep{nosofsky:gcm,rosch,Corter:1992,Murphy:1985}.  Empirical
studies of category acquisition and representation, have been
predominantly based on small-scale laboratory
experiments. In a typical experiment, human subjects are presented with 
small sets of often artificial concepts, such as binary 
strings~\cite{Anderson:1991} or colored shapes~\cite{Lee:2002}, with strictly 
controlled features~\citep{bornstein:online,medinschaffer,kruschke:93}. 
Hypotheses and principles of human categorization are established based on the 
processes and characteristics of the categorizations produced by the 
participants. The distribution of subjects participating in such studies is 
often skewed towards members of cultural and socioeconomic groups which are 
prevalent in the environment where the research is conducted, and typically 
consists to a large proportion of western, educated, wealthy and 
English-speaking participants often sampled from the even more specific 
population of college students.
The demographic and socioeconomic bias has been long recognized, and the 
question of how this bias might impact conclusions about human cognition in 
general~\citep{Henrich:2010} and category learning specifically is under 
active debate~\citep{Medin:2007}. Although
laboratory studies are invaluable for understanding categorization
phenomena in a controlled environment, they are also expensive and
time-consuming to conduct, and consequently problematic to scale.

\begin{figure}
 \begin{center}
\includegraphics[width=0.88\textwidth]{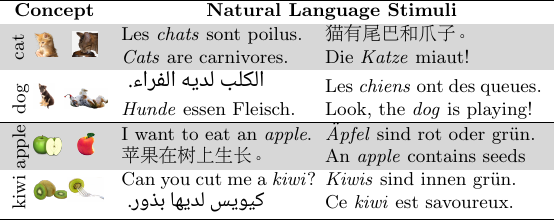}
\caption{\label{intro-examples}Illustration of model stimuli for five
  languages. Each stimulus contains a mention of a concept
  (e.g.,~\concept{cat} or \concept{apple}) in its local linguistic
  context. Concepts are grouped into categories
  (e.g.,~\category{animal} or \category{fruit}) based on the
  similarity of the contexts they occur in.}
 \end{center}
\end{figure}

In this work, we scale the investigation of category learning and 
representation along two axes: (1)~the complexity of the learning environment, 
and consequently the richness of learnable concept and category 
representations, and (2)~the diversity of languages and cultures considered in 
evaluation. We present a novel knowledge-lean, cognitively motivated Bayesian 
model which learns categories and their structured features jointly from large 
natural language text corpora in five diverse languages: Arabic, Chinese, 
English, French, and German. We approximate the learning environment using 
large corpora of natural language text.  Language has been shown to redundantly 
encode much of the non-linguistic information in the natural 
environment~\citep{Riordan:2011}, and to influence the emergence of 
categories~\citep{Gopnik:1987,Waxman:1995}. Besides text corpora can cover 
arbitrarily semantically complex domains, and are available across languages, 
providing an ideal test environment for studying categorization at scale.

\begin{figure}
\centering
	\includegraphics[width=1\linewidth]{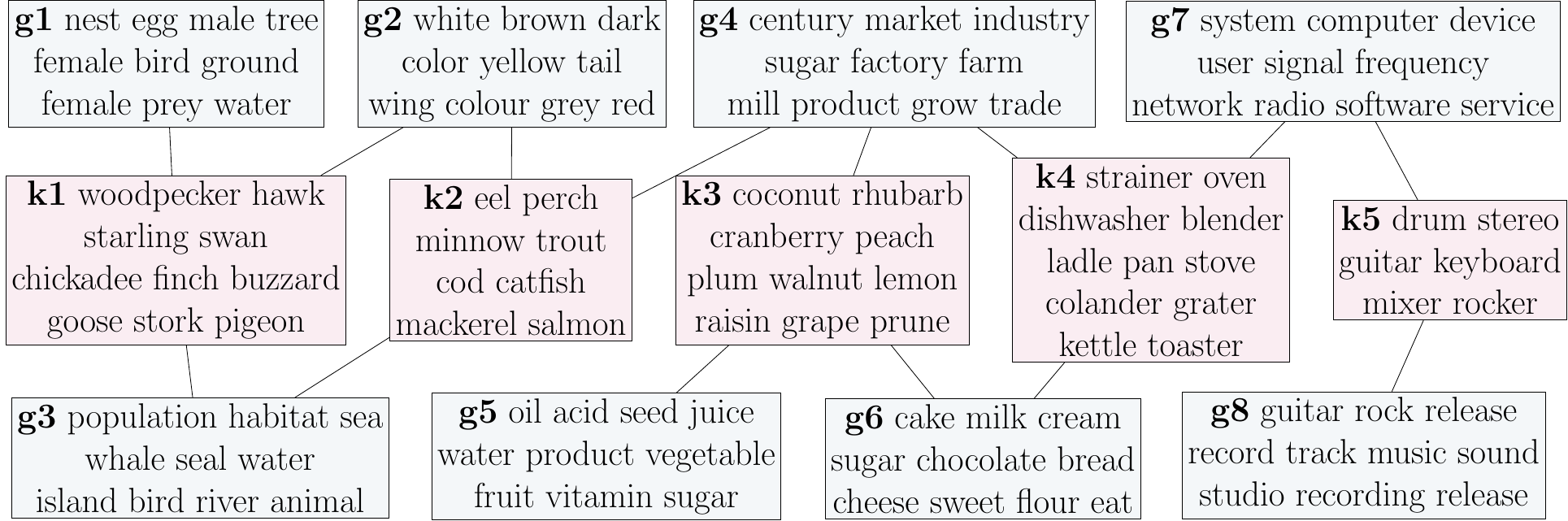}
        \caption{Examples of categories (light red) and feature types
          (light blue) inferred by BCF from the English
          Wikipedia. Connecting lines indicate associations learned
          between the category and the respective feature type. For
          example, category \category{fruit} ({\bf k3}) is associated
          with feature types \feature{trade} ({\bf g4})
          \feature{nutrition} ({\bf g5}), and \feature{processed food}
          ({\bf g6}).}
\label{fig:bcfenglish}
\end{figure}

Figure~\ref{intro-examples} illustrates example input to our model, and 
Figure~\ref{fig:bcfenglish} shows example categories and associated features as 
induced by our model from the English Wikipedia. Following prior 
work~\cite{Fountain:2011,Frermann:2016cogsci}, we create language-specific sets 
of stimuli, each consisting of a mention of target concept\footnote{Throughout 
this article we user the term {\it concept} to refer to Rosch's~\cite{rosch} 
basic-level categories, and the term {\it category} to refer to 
superordinate-level categories. (Superordinate-level) categories are inferred 
based on the features observed with observations of (basic-level) concepts.}
(e.g.,~\concept{apple}), within its local linguistic context (e.g.,~\{contains, 
seeds\}; cf.,~Figure~\ref{intro-examples}). We consider each stimulus an 
observation of the concept, i.e.,~the 
word referring to the concept is an instance of the concept itself, and its 
context words are a representation of its features. Our model infers categories 
as groups of concepts occurring with similar features; and it infers feature 
types as groups of features which co-occur with each other. The output of 
our model (cf.,~Figure~\ref{fig:bcfenglish}) are categories as clusters of 
concepts, each associated with a set of feature types, i.e.,~thematically 
coherent groups of features. We train a {\it separate} model on each of our 
target languages, each time presenting the model with input stimuli from the 
relevant language.

Computational models in general, and Bayesian models in particular, allow to 
investigate hypotheses about cognitive phenomena by systematically modifying 
the learning mechanism or available input while observing the learning outcome. 
Bayesian models have been applied to a variety of cognitive 
phenomena~\citep{Griffiths:2008,Chater:2010,Perfors:2011}, and category 
acquisition is no exception. Following from 
Anderson's~\citep{Anderson:1991,Sanborn:2010,Xu:2007} seminal work, a number of 
models have been developed, and tested in their ability to reproduce human 
behavior in laboratory settings by exposing the models to small sets of 
controlled inputs with restricted features. In this work we draw on the full 
potential of computational modeling by exposing our models to (a) more complex 
data reflecting the diversity of contexts in which concepts can be observed; 
and (b) a input data in different languages, shedding light on the 
applicability of computational cognitive models beyond the prevalent English 
test language. 

Categorization tasks in a laboratory environment typically involve stimuli with 
a small set of features which are {\it relevant} to the categorization target, 
eliminating the need to detect features, and discriminate them in their 
relevance. In the real world, however, concepts are observed in contexts and a 
substantial part of acquiring categorical knowledge involves learning which 
features are useful to discriminate among concepts. In fact, research has shown 
that humans learn 
features jointly with categories~\citep{Goldstone:2001,Schyns:1997} and that 
these features are themselves structured so as to represent the diversity and 
complexity of the properties exhibited in the 
world~\citep{Ahn:1998,McRae:2005,Spalding:2000}. Our novel model of category 
learning presented in this article, {\it jointly} learns categories 
and their {\it structured} features from large sets of informationally rich 
data.

Our work exemplifies the opportunities that arise from computational models and 
large data sets for investigating the mechanisms with which conceptual 
representations emerge, as well as the representations themselves in a broader 
context.  We simulate the acquisition of categories comprising hundreds of 
concepts by approximating the learning environment with natural language
text. Language has been shown to redundantly encode much of the non-linguistic 
information in the natural environment~\citep{Riordan:2011}, as well as 
human-like biases~\cite{Caliskan:2017}, and to influence the emergence of
categories~\citep{Gopnik:1987,Waxman:1995}. Text corpora are a prime
example of naturally occurring large-scale data 
sets~\citep{Goldstone:2016,Jones:2016,Paxton:2017}. In analogy to
real-world situations, they encapsulate rich, diverse, and potentially
noisy, information. The wide availability of corpora allows us to
train and evaluate cognitive models on data from diverse languages and
cultures.  We test our model on corpora from five languages, derived
from the online encyclopedia Wikipedia in Arabic, Chinese, French,
English, and German. Wikipedia is a valuable resource for our study
because it (a)~discusses concepts and their properties explicitly and
can thus serve as a proxy for the environment speakers of a language
are exposed to; and (b)~allows us to construct corpora which are
highly comparable in their content across languages, controlling for
effects of genre or style.

We present a series of evaluations investigating the quality of the induced 
categories and features. Leveraging a reference comprising hundreds of concepts 
and more than 30~categories, we demonstrate that our model learns meaningful 
categories in all five target languages. We furthermore show, through 
crowd-sourced evaluations involving native speakers of each target language, 
that the induced feature types are (a)~each thematically coherent and 
interpretable; and (b)~are associated with categories in comprehensible ways. 
We discuss language-specific idiosyncrasies emerging from the induced 
representations.

In the remainder of this article, we first review related literature,
before we present a cognitively motivated model for learning
categories and their structured representations from large natural
language corpora. We then evaluate the quality of the emerging
representations, as well as the generalizability of our model across
languages. Note that the primary goal of this work is not to {\it characterize}
differences in categories and features arising from different languages (even 
though this would be an interesting avenue for future work). Rather, we aim to 
demonstrate the utility of large-scale naturalistic datasets for cognitive 
modeling, and to verify mechanisms of categorization known from laboratory 
studies at scale and across communities.

\section{Related Work}
\label{rel-work}
In this work we leverage large-scale computational simulations to
advance our understanding of categories and features across languages
and cultures. Our research touches on the representation of
categories, concepts, and their features; the mechanisms with which
these are learnt; and the use of computational models and large-scale
naturalistic data sets to investigate these questions.




\subsection{Feature Representations of Concepts and Categories}
\label{4:litrev:structure}
Even though much empirical research glosses over this observation,
there is strong evidence that human conceptual representations are
structured~(see~\cite{Rips:2012} for a recent critique and overview of
cognitive studies of categorization). Categories mentally represent
the complex structure of the environment. They allow to make
inferences about concepts or categories that go beyond their perceived
similarities capturing abstract and potentially complex properties
(for example, the \feature{nutritional value} of \category{food} items,
or the \feature{emotional benefits} of \category{pets}). Much research
on human categorization is based on laboratory experiments where
subjects are presented with artificial stimuli represented by a
restricted set of task-relevant features. Observations of
\leaemph{natural} concepts, however, are often noisy or incomplete so
that a notion of systematic relations among features might be more
important here than under artificial conditions in the
lab~\citep{Malt:1984}.

The existence of structured features has received support through
behavioral results from a variety of categorization related tasks,
such as typicality rating~\citep{Malt:1984} or category-based
inductive inference~\citep{Heit:1994, Spalding:2000}. Experimental
evidence suggests that categories which on the surface do not seem to
contain a coherent set of members (e.g.,~the category~\category{pets})
are represented by an underlying set of abstract features which
explain the coherence of the category (e.g.,~\{keeps\_company,
lives\_in\_the\_house\}).
Varying the types of available features (e.g.,~providing
\feature{functional} information in addition to objects'
\feature{appearance}) leads to different categorization behavior both
in adults~\citep{Heit:1994} and
children~\citep{Traeuble:2007,jones:ea:91}, and different feature
types vary in their predictive value across categories. For example,
2-4-year old children categorize \category{food} items based on their
\feature{color}, however, \category{toys} are classified based on
their \feature{shape}~\cite{Macario:1991}. 

The structured nature of category features manifests itself in feature
norms. Feature norms are verbalized lists of properties that humans
associate with a particular concept~\citep{McRae:2005}. Features
collected in norming studies naturally fall into different types such
as \feature{behavior}, \feature{appearance} or
\feature{function}. This suggests that structure also emerges from
verbalized representations of concepts and features such as mentions
in natural language corpora, used as stimuli in this work. McRae et
al.~\cite{McRae:2005} collected a large set of feature norms for more
than 500~concepts in a multi-year study, and classified these using a
variety of theoretically motivated schemata, including the feature
type classification scheme developed in~\cite{Wu:2009}
and~\cite{Barsalou:1999}. Their work puts forward the hypothesis that
humans perform a ``mental simulation'' when describing a concept,
scanning the mental image they create as well as situations associated
with that image, and then verbalize it when producing features.



The model we present in this article aims to capture the evidence
summarized above, and represent categories as \leaemph{structured}
sets of features with varying degrees of association. Category-specific features are structured
into types which relate to a particular kind of property of a category
(e.g.,~the \feature{behavior} of \category{animals}). We also capture
the observation that features are defining for \emph{different} categories to
a varying degree \citep{Keil:1989, McRae:2002} in terms of
category-feature type associations (e.g.,~the feature
\feature{function} is highly defining for (or associated with) the
category \category{artifact}, but not for the category \category{animal}).

\subsection{Joint Learning of Categories and their Features}
\label{4:litrev:joint}
Although the majority of models of categorization assume a fixed set of features underlying the category acquisition and categorization process, there is 
increasing evidence that ``[...] a significant part of learning a category involves learning the features entering its representations.'' \cite[p.~681]{Schyns:1997}. Experimental evidence suggests that not only do features underly the categorization process but features themselves are susceptible to change over time and can be modified by the categories which emerge. Evidence ranges from changing featural perception as a result of expert education 
(e.g., wine tasters or doctors learning to interpret X-ray images) to neurological evidence revealing enhanced neural activity in experts when presented 
with pictures of their area of expertise~(see~\cite{Goldstone:2008} for an overview).

The influence of category learning on the perception and use of
features has been studied extensively using visual stimuli of varying
degrees of naturalness and familiarity. Experiments with drawings of
2-dimensional line segments~\cite{Pevtzow:1994} show that participants
who were exposed to categorization training prior to a feature
identification task identified the presence of category-defining
features faster than participants without prior training.  When asked
to categorize pictures of (systematically manipulated) human faces,
participants showed higher sensitivity to features relevant for the
categorization task \cite{Goldstone:2001,Goldstone:2001a}.



To the best of our knowledge, we present the first computational
investigation in the \leaemph{joint} emergence of categories and
features from large sets naturalistic input data.

\subsection{Computational Models of Category and Feature Induction}
\label{bcf:litrev:systems}
The tasks of category formation and feature learning have been
considered largely independently in the context of computational
cognitive modeling. Bayesian categorization models pioneered by
Anderson~\cite{Anderson:1991} and recently re-formalized by Sanborn et
al.~\cite{Sanborn:2006} aim to replicate human behavior in
small scale category acquisition studies, where a fixed set of simple
(e.g.,~binary) features is assumed. Informative features are pre-defined and available to the model. The BayesCat
model~\citep{Frermann:2015} is similar in spirit, but was applied to
large-scale corpora, while investigating incremental learning in the
context of child category acquisition (see also \cite{Fountain:2011}
for a non-Bayesian approach). BayesCat associates sets of features (context words)
with categories as a by-product of the learning process, however these
feature sets are independent across categories and are not optimized
during learning.

A variety of cognitively motivated Bayesian models have been proposed
for the acquisition of complex domain knowledge. Shafto et
al. \cite{Shafto:2011} present a joint model of category and feature
acquisition in the context of cross-categorization,~i.e.,~the
phenomenon that concepts are simultaneously organized into several
categorizations and the particular category (and features) that are
relevant depend on the context (e.g.,~concepts of the category
\category{food} can be organized based on their \feature{nutritional}
or \feature{perceptual} properties).
However, while \cite{Shafto:2011} present their model with category-specific data sets tailored towards their learning objective, we are interested in \leaemph{acquiring} categories and structured associated features jointly from thematically unconstrained corpora of natural text.

Another line of work~\citep{Kemp:2003, Perfors:2005} models the joint
learning of relevant features and domain-specific feature type biases
in children. They focus on the acquisition of domain-specific
representational structures (such as~hierarchies or clusters) and
discuss results in the context of word learning. In contrast to our
work, their model assumes a priori established categories (such as
\category{food} and \category{animals}), and learns from task-specific
data representations in the form of objects described by a limited set
of relevant features (even though a weighting of those features is
learnt). Perfors et al. \cite{Perfors:2009} present a Bayesian model
which simultaneously learns categories (i.e.,~groupings of concepts
based on shared features) and \leaemph{learns to learn} categories
(i.e.,~abstract knowledge about kinds of featural regularities that
characterize a category). They compare their model predictions against
behavioral data from adult participants, which limits the scope of
their experiments to small data sets. 

The ability to automatically extract feature-like information for
concepts from text would facilitate the laborious process of feature
norming,~i.e.,~eliciting features associated with concepts verbally
from human annotators~\cite{McRae:2005}, and improve the coverage of
concepts and their features. A few approaches to feature learning from
textual corpora exist, and they have primarily focused on emulating or
complementing norming studies by automatically extracting norm-like
properties from corpora (e.g., \concept{elephant} \feature{has-trunk},
\concept{scissors} \feature{used-for-cutting}). Steyvers
\cite{Steyvers:2010} uses a flavor of topic models to augment data
sets of human-produced feature norms. While vanilla topic models
\citep{Blei:2003} represent documents as sets of corpus-induced
topics, \cite{Steyvers:2010} additionally use topics derived from
feature norms. The learnt topics yield useful extensions of the
original feature norms, with properties that were previously not
covered, suggesting that corpora are an appropriate resource for
augmenting feature norms of concepts.

Another line of research concerns text-based feature extraction. A
common theme in this line of work is the use of pre-defined syntactic
patterns \citep{Baroni:2010}, or manually created rules specifying
possible connection paths of concepts to features in dependency trees
\citep{Devereux:2009, Kelly:2014}. While the set of syntactic patterns
pre-defines the relation types the system can capture, the latter
approach can extract features which are a priori unlimited in their
relation to the target concept. Once extracted, the features are
typically weighted using statistical measures of association in order
to filter out noisy instances. Similar to our own work, the motivation
underlying these models is large-scale unsupervised feature extraction
from text. These systems are not cognitively motivated
\leaemph{acquisition} models, however, due to (a)~the assumption of
involved prior knowledge (such as syntactic parses or manually defined
patterns), and (b)~the two stage extraction-and-filtering process
which they adopt. Humans arguably do not first learn a large set of
potential features for concepts, before they infer their
relevance. The systems discussed above learn features for individual
concepts rather than categories.


To our knowledge, we propose the first Bayesian model that jointly
learns categories and their features from large sets of naturalistic
input data. Our model is knowledge-lean, it learns from raw text in a
single process, without relying on parsing resources, manually crafted
rule patterns, or post-processing steps; it is more plausible from a
cognitive point of view, and language agnostic. We present simulations 
with the same model on several languages varying in word order,
morphology, and phonology.


\section{Category and Feature Learning at Scale}
Computational models as simulators of cognitive processes have been
used successfully to shed light on a wide variety of
phenomena~\citep{{Chater:2010}}, including language
acquisition~\citep{Xu:2007}, generalization, and
reasoning~\citep{Griffiths:2006}. Bayesian models in particular are
amenable towards this goal, because they allow the modeler to
formalize hypotheses rigorously through sets of random variables and
their relations. They use the principled rules of Bayesian probability
to select ``good'' models which explain the observed data well. We
present a Bayesian model to investigate cognitive {\it processes} of
categorization, in correspondence to Marr's~\citep{Marr:1982} {\it
  computational} level of analysis, i.e., abstracting away from the
algorithms and biological substrates in which these processes are
situated.  Starting from Anderson's~\citep{Anderson:1991} pioneering
work on rational models of categorization, a variety of models, both
Bayesian~\citep{Sanborn:2006,Shafto:2011,Frermann:2016cogsci} and
non-Bayesian~\citep{kruschke:93,Fountain:2011} have been proposed.
Our work advances prior research by investigating for the first time
joint category and feature learning from noisy stimuli, across diverse
languages.

We present BCF, a cognitively motivated {\bf B}ayesian model for
learning {\bf C}ategories and structured {\bf F}eatures from large
sets of concept mentions and their linguistic contexts
(see~Figure~\ref{intro-examples}). Our model induces categories (as
groups of concepts), feature types which are shared across categories
(as groups of features or context words), and category-feature type
associations. Figure~\ref{fig:bcfenglish} shows example output of BCF
as learnt from the English Wikipedia, and
Figure~\ref{fig:multilingual} shows example categories and features
learnt for five additional languages.

BCF is a statistical Bayesian model. Given a large set of stimuli, it
learns {\it meaningful} categories and features from a countably
infinite set of all possible categorizations and representations. The
probability (or `meaningfulness') of any hypothetical categorization
and representation $h$ under the stimuli data $d$ can be evaluated
using Bayes' rule:

\begin{equation}
 p(h|d) \propto p(d|h) p(h),
\label{eqn:bayes}
\end{equation}
where $p(h)$ is the prior probability of $h$ under the specified model
and its assumptions; and $p(d|h)$ is the {\it likelihood} to observe
 data~$d$ given that hypothesis $h$ holds.

\subsection{The BCF Model}
BCF learns from an input corpus which consists of stimuli covering~ 
$\mathcal{L}$ target concepts, where the set of target concepts is specified by 
the modeler a priori. The model induces a categorization of these target 
concepts into $K$~categories; as well as a characterization of each category in 
terms of $G$ different feature types pertaining to different relevant 
properties. The number of categories, $K$, and the number of feature types, 
$G$, are model parameters. 

\begin{table}
 \begin{center}
  \begin{tabular}{cll}
  \hline
   {\bf symbol}   & \multicolumn{2}{c}{\bf explanation} \\ \hline
   $d\in\{1..D\}$ & stimulus                    & (e.g.,~{``This {dog} likes to catch balls.''})\\
   $c\in\{1..C\}$ & \multicolumn{2}{l}{concept mention in stimulus  (e.g.,~{``dog''})} \\
   $i\in\{1..I\}$ & \multicolumn{2}{l}{context word positions in stimulus} \\\hline
   $\ell\in\{1..\mathcal{L}\}$ & concept types  & (e.g.,~\concept{cat, dog, chair, table})\\
   $f\in\{1..V\}$ & features                    & (e.g.,~\{runs, barks, eats, red, made\_of\_wood\})\\
   $k\in\{1..K\}$ & categories                  & (e.g.,~\category{animal, furniture})\\
   $g\in\{1..G\}$ & feature types               & (e.g.,~\feature{behavior, appearance})\\\hline
   $\theta$       & \multicolumn{2}{l}{$K$-dimensional parameter vector of category distribution}\\
   $\{\mu_{k}\}_{k=1}^{K}$      & \multicolumn{2}{l}{$G$-dimensional parameter vectors of feature type distributions}\\
   $\{\phi_g\}_{g=1}^G$       & \multicolumn{2}{l}{$V$-dimensional parameter vectors of word distributions}\\\hline
  \end{tabular}
  \caption{Notational overview of the BCF model (the category and feature type labels are provided for illustration; BCF is an unsupervised clustering 
model which induces unlabeled categories and feature types).}
  \label{bcf:notation}
 \end{center}
\end{table}

\begin{figure}
 \begin{subfigure}{\textwidth}
 \caption{\label{gen-story}Generative story of BCF.}
 \begin{algorithmic}
 \State Generate category distribution, $\hspace{2ex}\theta \sim Dir(\alpha)$ \vspace{1ex}
 \For{concept type $\ell=1..\mathcal{L}$}
   \State Generate category, $\hspace{2ex}k^\ell \sim Mult(\theta)$
 \EndFor \vspace{1ex}
 \For{category $k=1..K$}
   \State Generate feature type distribution, $\hspace{2ex}\mu_k \sim Dir(\beta)$
 \EndFor \vspace{1ex}
 \For{feature type $g=1..G$}
   \State Generate feature distribution, $\hspace{2ex}\phi_g \sim Dir(\gamma)$
 \EndFor \vspace{1ex}
  \For{stimulus $d=1..D$}
  \State Observe concept $c^d$ and retrieve category $k^{c^d}$
  \State Generate a feature type, $\hspace{2ex}g^d \sim Mult(\mu_{k^{c^d}})$\vspace{0.5ex}
    \For{feature position $i=1..I$}
      \State Generate a feature $\hspace{2ex}f_{d,i} \sim Mult(\phi_{g^d})$ 
    \EndFor
  \EndFor 
  \end{algorithmic}
\end{subfigure}
\vspace{5ex}

\begin{subfigure}{\textwidth}
\caption{\label{bcf:plate-diagram}Plate diagram of BCF.}
 \begin{center}
     \scalebox{1.5}{  \begin{tikzpicture}
 
  \node[latent]                (G)     {g} ; %
  \node[obs, left=of G]       (Wc)    {c} ; %
  \node[obs, right=of G]      (Wf)    {f} ; %
  \node[latent, below=1.5cm of Wf]      (k)    {$k^\ell$} ; %
  \node[latent, right=of k]   (theta) {$\theta$};
  \node[latent, dashed, right=of theta]   (alpha) {$\alpha$};
  \node[latent, below=1.5cm of G]     (mu)    {$\mu_{k}$}; %
  \node[latent, dashed, left=of mu]     (beta) {$\beta$};
  \node[latent, right=of Wf]     (phi) {$\phi$};
  \node[latent, dashed, right=of phi]     (gamma) {$\gamma$};

  \draw[->] (Wc) -- (G);
  \draw[->] (mu) -- (G);
  \draw[->] (G) -- (Wf);
  \draw[->] (beta) -- (mu);
  \draw[->] (k) -- (G);
  \draw[->] (theta) -- (k);
  \draw[->] (alpha) -- (theta);
  \draw[->] (phi) -- (Wf);
  \draw[->] (gamma) -- (phi);
   
  \plate {plate2} {
      (Wf)
  } {$I$};
  \plate {plate1} { %
    (Wc)(G) (plate2)%
  } {$D$}; %
  \plate {plate3} {
      (k)
  } {$\mathcal{L}$};
  \plate {plate4} {
      (mu)
  } {$K$};
  \plate {plate5} {
      (phi)
  } {$G$};
   
  \end{tikzpicture}}  

 \end{center}
\end{subfigure}
\caption{Top (a): The generative story of the BCF model. Observations~$f$ and
  latent labels~$k$ and~$g$ are drawn from Multinomial distributions
  ($Mult$). Parameters for the multinomial distributions are drawn
  from Dirichlet distributions ($Dir$). Bottom (b): The plate diagram of the BCF model. Shaded nodes indicate observed variables, clear nodes denote latent 
variables, and dashed nodes indicate constant
hyperparameters.}
\end{figure}
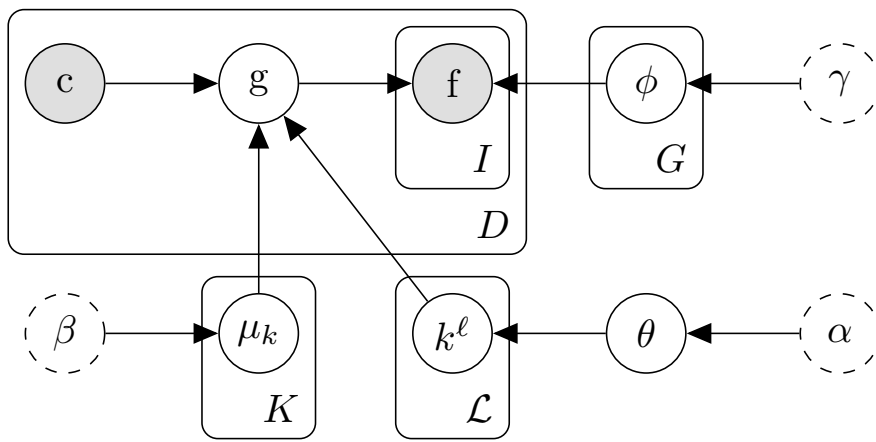


A notational overview is provided in Table~\ref{bcf:notation}. The generative 
story of our model is displayed in Figure~\ref{gen-story}, and Figure 
\ref{bcf:plate-diagram} shows the plate diagram representation of BCF. The 
generative story proceeds as follows. We assume a global multinomial 
distribution over categories~$Mult(\theta)$. Its parameter vector~$\theta$ is 
drawn from a symmetric Dirichlet distribution with hyperparameter~$\alpha$. For 
each target concept~$\ell=[1...\mathcal{L}]$, we draw a category~$k^\ell$ 
from~$Mult(\theta)$. For each category~$k$, we draw an independent set of 
multinomial parameters over feature types,~$\mu_k$, from a symmetric 
Dirichlet distribution with hyperparameter~$\beta$, reflecting the relative 
relevance of each feature type towards this category. Finally, we associate 
each feature type with representative words from our feature vocabulary 
$f\in1...V$, by drawing a multinomial distribution over 
features,~$Mult(\phi_g)$, from a symmetric Dirichlet distribution with 
hyperparameter~$\gamma$. From this set of global representations, we can 
generate sets of stimuli~$d=[1...D]$ as follows: we first retrieve the 
category~$k^{c^d}$ of an observed concept~$c^d$; we then generate a feature 
type~$g^d$ from the category's feature type distribution~$Mult(\mu_{k^{c^d}})$; 
and finally, for each context position~$i=[1...I]$ we generate 
feature~$f_{d,i}$ from the feature type-specific feature 
distribution~$Mult(\phi_{g^d})$.

According to the generative story outlined above, the joint probability of the 
model over latent categories, latent feature types, model parameters, and data 
factorizes as:
\begin{equation}
\begin{aligned}
 p&(g,f,\mu,\phi,\theta,k|c,\alpha,\beta,\gamma) = \\
    &p(\theta|\alpha)\prod_{\ell}p(k^\ell|\theta)\prod_kp(\mu_k|\beta)\prod_gp(\phi_g|\gamma)\prod_dp(g^d|\mu_{k^{c^d}})\prod_ip(f^{d,i}|\phi_{g^d}).
\end{aligned}
\end{equation}
Since we use conjugate priors throughout, we can integrate out the model parameters analytically, and perform inference only over the latent variables, namely 
the category and feature type labels associated with the stimuli.

In sum, our model takes as input a text corpus of concept mentions in
their local context, and infers a concept categorization, a global set
of feature types, as well as a distribution over feature types per
category. After integrating out model parameters where possible, we
infer two sets of latent variables:
\begin{itemize}
 \item[(1)] feature type-assignments to each stimulus $\{g\}^D$,
 \item[(2)] category-assignments to each concept type $\{k\}^\mathcal{L}$.
\end{itemize}

The next section introduces a learning algorithm in the form of a
Gibbs sampler for approximate estimation of these parameters.

\subsection{Approximate Inference for BCF}
\label{bcf:sec:gibbs}
\begin{algorithm}[t!]
 \begin{algorithmic}[1]
   \State{Input: model with randomly initialized parameters.}
   \State{Output: posterior estimate of $\boldsymbol\theta, \boldsymbol\phi$, and $\boldsymbol\mu$.}
   \Repeat{}
   \For{stimulus $d$} \Comment{Sample stimulus-feature type assignments}
      \State - \text{decrement stimulus $d$-related counts} 
       \State - Sample \hspace{1ex} $g^d \sim 
p(g_{k^{c^d}}^d=i|\mathbf{g}_{k^{c^d}}^{-d}, \mathbf{f}^{-}, k^{c^d}, \beta, 
\gamma)$ 
\text{\hspace{2ex}Equation~(\ref{bcf:posterior-featuretype-simplified})}
      \State - \text{update stimulus $d$-related counts}
   \EndFor
   \For{concept $c$} \Comment{Sample concept-category assignments}
     \State -  retrieve category $k^c$
       \State - \text{decrement concept $c$-related counts} 
       \State -  Sample \hspace{1ex}  $k^c \sim p(k^\ell=j|\mathbf{g}_{k^\ell}, 
\mathbf{k^{-}}, \alpha, \beta)$
\text{\hspace{2ex}Equation~(\ref{bcf:posterior-category-simplified})}
       \State - \text{update concept $c$-related counts}
    \EndFor
   \Until{convergence}
 \end{algorithmic}
 \caption{The Gibbs sampling algorithm for the BCF model.}
 \label{bcf:gibbs-algo}
\end{algorithm}

Exact inference in the BCF model is intractable, so we turn to approximate posterior inference to discover the distribution over value assignments to latent variables given the observed data. In this section we introduce a Gibbs sampling algorithm~\citep{Geman:1984,Bishop:2006} which is a Markov chain Monte Carlo algorithm which iteratively computes values of individual random variables in the model, based on the current value assignments of all other random variables. The sampling procedure for BCF is summarized in Algorithm~\ref{bcf:gibbs-algo}. The Gibbs sampler repeatedly iterates over the training corpus and resamples values of the latent variables. One Gibbs iteration for our model consists of two blocks:

\paragraph{Resampling stimulus-feature type assignments} In the
first block we iterate over input stimuli~$d$, and resample each
stimulus-feature type assignment~$g^d$ from its full conditional
posterior distribution over feature types conditioned on (a)~the
values assigned to all other latent variables unrelated to the current
variable of interest,~i.e,~all features except those in stimulus~$d$,
$\big(\mathbf{f}^{-}\big)$, and all stimulus-feature type assignments
except the one to stimulus~$d$, $\big(\mathbf{g}_{k^{c^d}}^{-d}\big)$;
(b)~the category currently assigned to~$d$'s target concept~$c$,
$\big(k^{c^d}\big)$; and (c)~the relevant hyperparameters~$\big(\beta,
\gamma\big)$:

\begin{eqnarray}
p(g_{k^{c^d}}^d=i&|&\mathbf{g}_{k^{c^d}}^{-d},\ \mathbf{f}^{-},\ k^{c^d}=j,\ 
\beta,\ \gamma) \\
\label{bcf:posterior-featuretype}
 &=&\ \ p(g_{k^{c^d}}^d=i|\mathbf{g}_{k^{c^d}}^{-d}, k^{c^d}=j,\beta) \times 
\hspace{2ex} p(f^d|\mathbf{f}^-, g_{k^{c^d}}^d=i, \gamma)\\
 \label{bcf:posterior-featuretype-simplified}
 &\propto&\ \ \frac{(n_i^j+\beta)}{(\sum_{i} n^j_i+\beta)} \times \hspace{2ex} 
  \frac{\prod_v\prod_{a=1}^{f_v}(n_v^{i}+\gamma+a)}{\prod_{a=1}^{f_*}(\sum_v n_v^{i}+\gamma+a)}.
\end{eqnarray}
The factorization of the posterior distribution in
(\ref{bcf:posterior-featuretype}) follows from the model structure as
described above and shown in the plate diagram in
Figure~\ref{bcf:plate-diagram}. The posterior distribution factorizes
into the probability of a particular feature type~$i$ and the
probability of the observed features in the stimulus given that
feature type. Because of the Dirichlet-Multinomial conjugacy in our
model, these two distributions can be straightforwardly computed using
only the counts of current value-assignments to all variables in the
model except the ones currently resampled
(equation~(\ref{bcf:posterior-featuretype-simplified})): the
probability of a hypothetical feature type~$i$ is proportional to the
number of times it has assigned previously to stimuli with observed
category~$j$, $n^j_i$, smoothed by the Dirichlet
parameter~$\beta$. Similarly, the probability of the observed features
of stimulus~$d$ under hypothetical feature type~$i$ is proportional to
the number of times each individual feature~$v$ in $d$ has been
observed under feature type~$i$, $n^i_v$ (smoothed by the Dirichlet
parameter~$\gamma$). In the second term in
(\ref{bcf:posterior-featuretype-simplified}), $f_v$ refers to the
count of any particular feature~$v$ in stimulus~$d$, and~$f_*$ refers
to the number of features in~$d$ (irrespective of their value).

We compute the (unnormalized) probabilities of individual hypothetical feature types~$i$ as explained above. These values are then normalized and a new feature type is sampled from the resulting distribution.

\paragraph{Resampling concept-category assignments} The second block of our Gibbs sampler performs a sweep over all concept types~$\ell \in 
\{1...\mathcal{L}\}$, and resamples each concept type~$\ell$'s category 
assignment~$k^{\ell}$. Similarly to the process described above, the new 
category assignment of concept~$\ell$ is resampled from its full conditional 
distribution over categories conditioned on (a)~all concept-category assignments 
except for~$k^\ell$, $\big(\mathbf{k^{-}}\big)$; (b)~the feature type 
assignments relevant to concept~$\ell$, $\big(\mathbf{g}_{k^\ell}^-\big)$; and 
(c)~all relevant hyperparameters~$\big(\alpha, \beta\big)$:

\begin{eqnarray}
 \label{bcf:posterior-category}
 p(k^\ell=j|\mathbf{g}_{k^\ell}^-,\ \mathbf{k^{-}},\ \alpha,\ \beta) 
  &=&p(k^\ell=j|\mathbf{k}^-,\alpha) \times\hspace{2ex} 
p(\mathbf{g}_{k^\ell}|\mathbf{g}^{-}_{k^\ell}, k^\ell=j, \beta)\\
  \label{bcf:posterior-category-simplified}
  &\propto&(n^j+\alpha) \times\hspace{2ex} 
\frac{\prod_g\prod_{a=1}^{f^\ell_g}(n^{j}_g+\beta+a)}{\prod_{a=1}^{f^\ell_*}
(\sum_gn^{j}_g+\beta+a)}.
\end{eqnarray}
Based on the independence assumptions in our model, this probability
factorizes into the prior probability of hypothetical category~$j$ and
the probability of feature types observed with concept~$\ell$ under
the hypothetical category~$j$
(equation~(\ref{bcf:posterior-category})). As above, these
probabilities can be computed purely based on counts of
variable-assignments in the current sampler state
(equation~(\ref{bcf:posterior-category-simplified})).  In the second
term of (\ref{bcf:posterior-category-simplified}), $f^\ell_g$ refers
to the number of times feature type~$g$ was assigned to a stimulus
containing concept type~$\ell$, and~$f^\ell_*$ to the number of
stimuli containing~$\ell$ (irrespective of the assigned feature type).

Using the procedure described above we compute an (unnormalized) probability for each hypothetical category, normalize the probabilities and resample concept $\ell$'s category~$k^\ell$ from the resulting distribution.


\section{Experimental Setup}
\label{sec:experimental-setup}

Can we simulate category acquisition from large amounts of textual
data using cognitively motivated computational models, and infer
meaningful representations across languages?  

We approach this question by applying BCF to data sets in five
languages: English, French, German, Arabic, and Chinese. We train five
models in total, one per language, each time using stimuli from the
respective language alone. We evaluate induced categories by
comparison against a human-created reference categorization; and
collect judgments on the coherence of learnt feature types, and their
relevance to their associated categories from large crowds of native
speakers.

Is the structure and architecture of BCF appropriate and necessary for
category and structured feature learning?  We answer this question by
comparing BCF against a variety of related models. First, we report a
random baseline which assigns concepts to categories at
random. Secondly, we compare against a model entirely based on word
co-occurrence. Unlike BCF, the co-occurrence model cannot learn
categories and features jointly, and has no notion of feature
structure. It uses $k$-means clustering~\citep{MacKay:2002} to group
concepts into categories, and, subsequently, group features into
feature types for each category (see Section~\ref{coocc-baseline}).
Finally, we compare BCF against BayesCat, a cognitively motivated
Bayesian model of category
acquisition~\citep{Frermann:2016cogsci}. Like BCF, it draws
inspiration from topic modeling, however, BayesCat does not learn
categories and features jointly, and does not acquire structured
feature representations.

In the following we describe our data set, as well as the set of
models we compare BCF against. Next, we present a series of
simulations evaluating the quality of the induced categories, their
features, and their relevance to the associated categories.

\subsection{Experimental Stimuli}
\label{sec:stimuli-corpora}
  \begin{table}[t]
\begin{center}
 \begin{tabular}{lrrrrr}
  \hline
                     & \multicolumn{1}{c}{\bf en} & \multicolumn{1}{c}{\bf ar} & 
\multicolumn{1}{c}{\bf ch} & \multicolumn{1}{c}{\bf fr} &\multicolumn{1}{c}{\bf 
ge} \\\hline
 {\bf concepts}      &491       &394       &450       &484       &482      \\
 {\bf features}      &5,898     &5,870     &6,516     &6,416     &6,981    \\
 {\bf stimuli}       &418,755   &86,908    &147,386   &258,499   &233,175  \\\hline
  \end{tabular}
  \caption{Datasets obtained from language-specific Wikipedia dumps
    for Arabic (ar), Chinese (ch), English, (en), French (fr), and
    German (ge). Number of stimuli (concept mentions in linguistic
    context), concepts, and features (context word types) are shown.}
    \label{tab:corpus:sizes}
\end{center}
\end{table}
Our simulations focused on 491 basic-level concepts of living and non-living things, taken from two
previous studies of concept
representation~\citep{McRae:2005,Vinson:2008}, for which we learn (a)
a categorization and (b) structured feature
representations. Human-created gold standard categorizations of the
concepts into 33 categories are
available~\citep{Vinson:2008,Fountain:2010}. Since the original
studies were conducted in English, we collected translations of the
target concepts and their categories into Arabic, Chinese, French, and
German, created by native speakers of the target language. The final
number of concepts differs across languages, because some English
concepts do not exist (or do not have the same translation) in the
target language. Concept sets and categorizations for all languages
were made available as part of this submission. 


We created language specific sets of input stimuli (as illustrated in
Figure~\ref{intro-examples}). For each target language we created a corpus as follows: We
used a subset of articles from the Linguatools Wikipedia
dump\footnote{\url{http://linguatools.org/tools/corpora/wikipedia-monolingual-corpora/}};
we tokenized, POS-tagged and lemmatized the corpus, and removed stopwords
using language-specific lists. From this data set we derived a set of
input stimuli as mentions of a concept from the reference set of concepts in sentence context (cf.,~Figure~\ref{intro-examples}). In order to obtain balanced data sets, we
automatically filtered words of low importance to a concept from contexts, using the
term-frequency-inverse-document-frequency (tf-idf) metric. After filtering, we only kept stimuli with $3 \leq n \leq 20$ context
words and at most 1,000 stimuli per target concept. Table~\ref{tab:corpus:sizes} summarizes the statistics of
the resulting data sets. The number of stimuli varies across languages
as a function of the number of target concepts, and the size of the
respective Wikipedia corpus.

\subsection{Comparison Models}
\label{sec:comparison-models}

We compared BCF against various models explained below. All
experiments follow the same experimental protocol, i.e., we train {\it
  separate} instances of the same model on each language.

\paragraph{Strudel} Following a pattern-based approach, Strudel
automatically extracts features for concepts from text collections. It
takes as input a part of speech-tagged corpus, a set of target
concepts and a set of 15~hand-crafted rules. Rules encode general, but
quite sophisticated linguistic patterns which plausibly connect nouns
to descriptive attributes (e.g., \leaemph{extract an adjective as a
  property of a target concept mention if the adjective follows the
  mention, and the set of tokens in between contain some form of the
  verb `to be'.}~\citep{Baroni:techreport}). Strudel obtains a large
set of concept-feature pairs by scanning the context of every
occurrence of a target concept in the input corpus, and extracting
context words that are linked to the target concept by one of the
rules. Each concept-feature pair is subsequently weighted with a
log-likelihood ratio expressing the pair's strength of
association. Baroni et al.~\cite{Baroni:2010} show that the learnt
representations can be used as a basis for various tasks such as
typicality rating, categorization, or clustering of features into
types. We obtained Strudel representations from the same Wikipedia
corpora used for extracting the input stimuli for BCF and
BayesCat. Note that Strudel, unlike the two Bayesian models, is not a
cognitively motivated \leaemph{acquisition} model, but a system
optimized with the aim of obtaining the best possible features from
data.

\paragraph{Co-occurrence Baseline} \label{coocc-baseline} Strudel relies on manually constructed linguistic patterns, and is
consequently not directly applicable across languages. We report a
baseline which is constructed to resemble Strudel, but does not rely
on linguistic features. It allows us to assess whether pure
co-occurrence counts provide a strong enough learning signal for
category and feature induction across languages. This model represents
each concept $c$ as a vector with dimensions corresponding to its
co-occurrence counts with features $f$ (i.e.,~context words), capped
by a minimum number of required observations, approximating the
concept-feature association:
\begin{equation}
 \label{cooc-assoc}
 assoc(c,f) = \mathcal{N}(c,f).
\end{equation}
We obtained categories by clustering concepts based on their vector
representations using $k$-means clustering~\citep{Lloyd:1982}. Based
on these categories, we obtained feature types by~(1) collecting all
features associated with at least half the concepts in the category;
and (2)~clustering these features into feature types using $k$-means
clustering.

\paragraph{BayesCat} Similar to BCF, BayesCat is a knowledge-lean {\it
  acquisition} model which can be straightforwardly applied to input
from different languages. It induces categories $z$ which are
represented through a distribution over target concepts $c$, $p(c|z)$,
and a distribution over features $f$ (i.e., individual context words),
$p(f|z)$. BayesCat, like BCF, is a Bayesian model and its parameters
are inferred using approximate MCMC inference, in the form of a Gibbs
sampler. Unlike BCF, however, BayesCat does not induce structured
feature representations, and comparing it to BCF allows us to evaluate
the advantage of {\it joint} category and feature learning. BayesCat
induces categories represented through unstructured bags-of-features.
As such, the model structure of BayesCat is closely related to topic
models such as Latent Dirichlet Allocation~(LDA;
\cite{Blei:2003}). Comparing our proposed model against BayesCat
allows us to shed light on the benefit of more sophisticated model
structure which allows to learn features jointly with categories,
compared to the information that can be captured in vanilla topic
models.  For our human evaluation in Section~\ref{exp-3} we construct
feature types from BayesCat features as follows. First we represent
each feature $f$ as its probability under each category
$p(z|f)$. Based on this representation, we again employ $k$-means
clustering to group features into $G$~global feature types
$g$. Finally, we compute category-featuretype associations as:

\begin{equation}
 p(g|z) = \sum_{f\in g} p(f|z),
\end{equation}
where $p(f|z)$ is learnt by BayesCat. 

While BCF induces a hard assignment of concepts to categories, BayesCat
learns a soft categorization. Soft assignments can be converted into hard
assignments by assigning each concept $c$ to its most probable
category $z$,

\begin{equation}
 z(c) = \max_z p(c|z) p(z|c).
\end{equation}

\paragraph{Model Parameters} 

Across all simulations we trained BCF to induce \mbox{$K=40$}
categories and $G=50$ feature types which are shared across
categories. We ran the Gibbs sampler for 1,000 iterations, and report
the final most likely representation. We trained BayesCat on the same
input stimuli as BCF, with the following parameters: the number of
categories was set to~$K=40$, and the hyperparameters to~$\alpha=0.7,
\beta=0.1$, and $\gamma=0.1$. From the learnt representations, we
induced $G=50$ global feature types as described above.  Again results
are reported as averages over 10~runs of 1,000~iterations of the Gibbs
sampler. The co-occurrence model induces \mbox{$K=40$} categories,
and, subsequently, \mbox{$G=5$} feature types for each category.

\section{Experiment 1: Category Quality}
In this simulation, We evaluate the extent to which model-induced
categories resemble the human created reference categorization.  We
report results on cluster quality for BCF, BayesCat, and the frequency
baseline for our five target languages.  For English, we additionally
report results for Strudel. We also lower-bound the performance of all
models with a random clustering baseline (random), which randomly
assigns all concepts to $K=40$ categories.

\subsection{Method}
The output clusters of an unsupervised learner do not have a natural 
interpretation. Cluster evaluation in this case involves mapping the induced 
clusters to a gold standard and measuring to what extent the two clusterings 
(induced and gold) agree \citep{Lang:2011}.  Purity ($pu$) measures the extent 
to which each induced category contains concepts that share the same gold 
category.  Let~$G_{j}$ denote the set of concepts belonging to the~$j$-th gold 
category and~$C_{i}$ the set of concepts belonging to the~$i$-th cluster.  
Purity is calculated as the member overlap between an induced category and its 
mapped gold category. The scores are aggregated across all induced 
categories~$i$, and normalized by the total number of category members~$N$:

\begin{equation}
  \text{pu} = \frac{1}{N} \sum_i \max\limits_j |C_i \cap G_j|
\end{equation}
Inversely, collocation ($co$) measures the extent to which
\leaemph{all}  members of a gold category are present in an induced
category. For each gold category we determine  the induced category
with the highest concept overlap and then compute the number of shared
concepts. Overlap
scores are aggregated over all gold categories~$j$, and normalized by
the total number of category members~$N$:

\begin{equation}
 \text{co} = \frac{1}{N} \sum_j \max\limits_i |C_i \cap G_j|
\end{equation}
Finally, the harmonic mean of purity and collocation can be used to
report a single measure of clustering quality.  If~$\beta$ is
greater than~1, purity is weighted more strongly in the calculation,
if~$\beta$ is less than~1, collocation is weighted more strongly:

\begin{equation}
  \text{F}_{\beta} = \frac{(1+\beta) \cdot pu \cdot co}{(\beta \cdot
    pu) +co}
\end{equation}
We additionally report results in terms of \leaemph{V-Measure} (VM,
\citep{Rosenberg:2007}) which is an information-theoretic measure. VM
is analogous to F-measure, in that it is defined as the weighted
harmonic mean of two values, \leaemph{homogeneity} (VH, the precision
analogue) and \leaemph{completeness} (VC, the recall analogue):

\begin{eqnarray}
  \text{VH} &=& 1-\frac{H(G|C)}{H(G)}\\
  \text{VC} &=& 1-\frac{H(C|G)}{H(C)}\\
  \text{VM} &=& 1-\frac{(1+\beta) \cdot VH \cdot VC}{(\beta \cdot VH)+VC}
\end{eqnarray}
where~$H(\cdot)$ is the entropy function;~$H(C|G)$~denotes the
conditional entropy of the induced class~$C$ given the gold standard
class~$G$ and quantifies the amount of additional information
contained in~$C$ with respect to~$G$.
The various entropy values involve the estimation of the joint
probability of classes~$C$ and~$G$:

\begin{equation}
\label{eq:non-fuzzy}
\hat{p}(C,G) = \frac{\mu(C \cap  G)}{N}
\end{equation} 

\subsection{Results}
\label{sim1-results}
\begin{table}[ht!]
\begin{center}
\begin{tabular}{clccccccc}
 \hline
& model               & \multicolumn{3}{c}{PCF1} & 
\multicolumn{3}{c}{V-Measure}\\
&                                        & $pu$ & $co$ & $F_1$ & VH & VC & 
VM\\\hline
\multirow{4}{*}{English}   & BCF      & 0.552 &0.432 &0.484 & 0.652 &0.598 
&0.623 & \\
                      & BayesCat & 0.551 &0.429 &0.482 & 0.646 &0.577 &0.609 &\\
                      & Strudel  & 0.572 &0.442 &{\bf 0.499} &0.662 &0.590 &{\bf 
0.624} \\\cdashline{2-8}
                      & co-occ   & 0.550 &0.394 &0.459 & 0.626 &0.559 &0.591 & 
\\
                      & random   & 0.193 &0.135 &0.159& 0.317 &0.282 
&0.298\\\hline
\multirow{3}{*}{German}   & BCF      & 0.454 &0.400 &{\bf 0.425} & 0.545 &0.523 
&0.534 & \\
                      & BayesCat & 0.458 &0.378 &0.414 & 0.563 &0.513 &{\bf 
0.537} &\\\cdashline{2-8}
                      & co-occ   & 0.338 &0.387 &0.361 & 0.408 &0.435 &0.421 & 
\\
                      & random   & 0.194 &0.134 &0.158& 0.316 &0.280 
&0.297\\\hline
\multirow{3}{*}{French}   & BCF      & 0.534 &0.441 &{\bf 0.483} & 0.632 &0.585 
&{\bf 0.608} &  \\
                      & BayesCat & 0.507 &0.415 &0.457 & 0.609 &0.558 
&0.582\\\cdashline{2-8}
                      & co-occ   & 0.459 &0.365 &0.407 & 0.544 &0.509 &0.526 & 
\\
                      & random   & 0.197 &0.134 &0.160& 0.319 &0.283 
&0.300\\\hline
\multirow{3}{*}{Chinese}   & BCF      & 0.441 &0.349 &{\bf 0.389} & 0.510 
&0.497 &0.503&  \\
                      & BayesCat & 0.430 &0.320 &0.367 & 0.532 &0.493 &{\bf 
0.512}\\\cdashline{2-8}
                      & co-occ   & 0.367 &0.327 &0.345 & 0.408 &0.422 &0.415& \\
                      & random   & 0.208 &0.135 &0.164& 0.325 &0.291 
&0.307\\\hline
\multirow{3}{*}{Arabic}   & BCF      & 0.408 &0.327 &{\bf 0.363} & 0.444 &0.446 
&0.445 \\
                      & BayesCat & 0.394 &0.298 &0.339 & 0.491 &0.462 &{\bf 
0.476}\\\cdashline{2-8}
                      & co-occ   & 0.261 &0.308 &0.283 & 0.312 &0.344 &0.327& \\
                      & random   &0.214 &0.125 &0.158& 0.329 &0.296 &0.312 
\\\hline
\end{tabular}
\caption{Quality of induced categories for BCF (this work), BayesCat,
  a co-occurrence baseline, Strudel (for English only), and a random baseline. 
Results are reported for English, German, French, Chinese and Arabic.}
\label{tab:categoryquality}
\end{center}
\end{table}
Table~\ref{tab:categoryquality} displays the results
for all five languages. BCF learns categories which most closely
resemble the human gold standard, and both BCF and the co-occurrence
model clearly outperform the random baseline. The Bayesian models, BCF and BayesCat, outperform the co-occurrence model across metrics and languages. For English, Strudel slightly outperforms BCF. Note, however, that, BCF \leaemph{learns} the categories from data, whereas for Strudel we construct the categories post-hoc after a highly informed feature extraction process (relying on syntactic patterns). It is therefore not surprising that Strudel performs well, and it is encouraging to see that BCF learns categories of comparable quality. 

\begin{figure*}[ht!]
    \begin{subfigure}[b]{\linewidth}
	\centering
	\includegraphics[width=\linewidth]{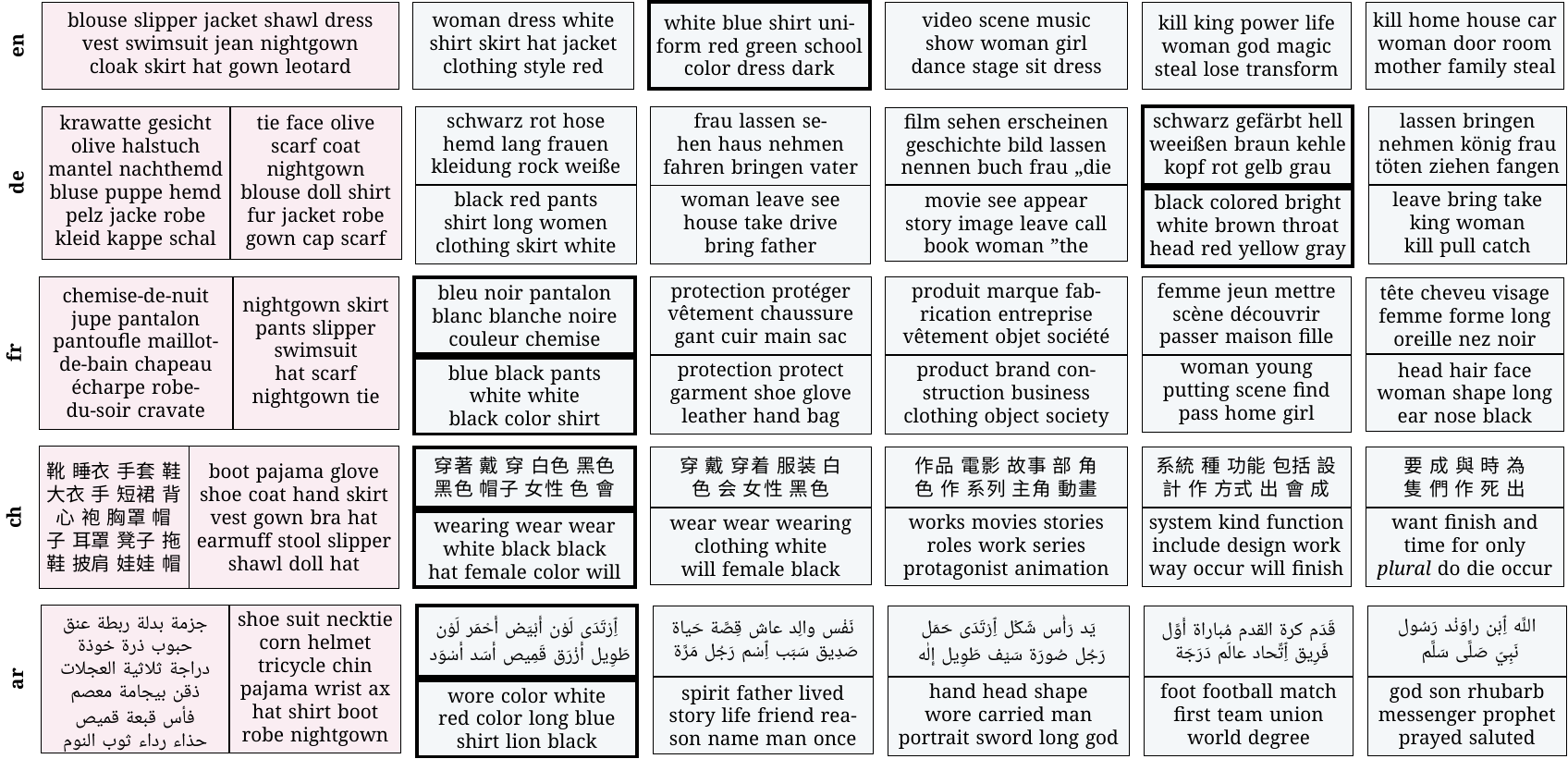}
	\caption{Category \category{clothing}}
        \label{fig:multilingual_clo}
    \end{subfigure}
    \begin{subfigure}[b]{\linewidth}
	\centering
	\includegraphics[width=\linewidth]{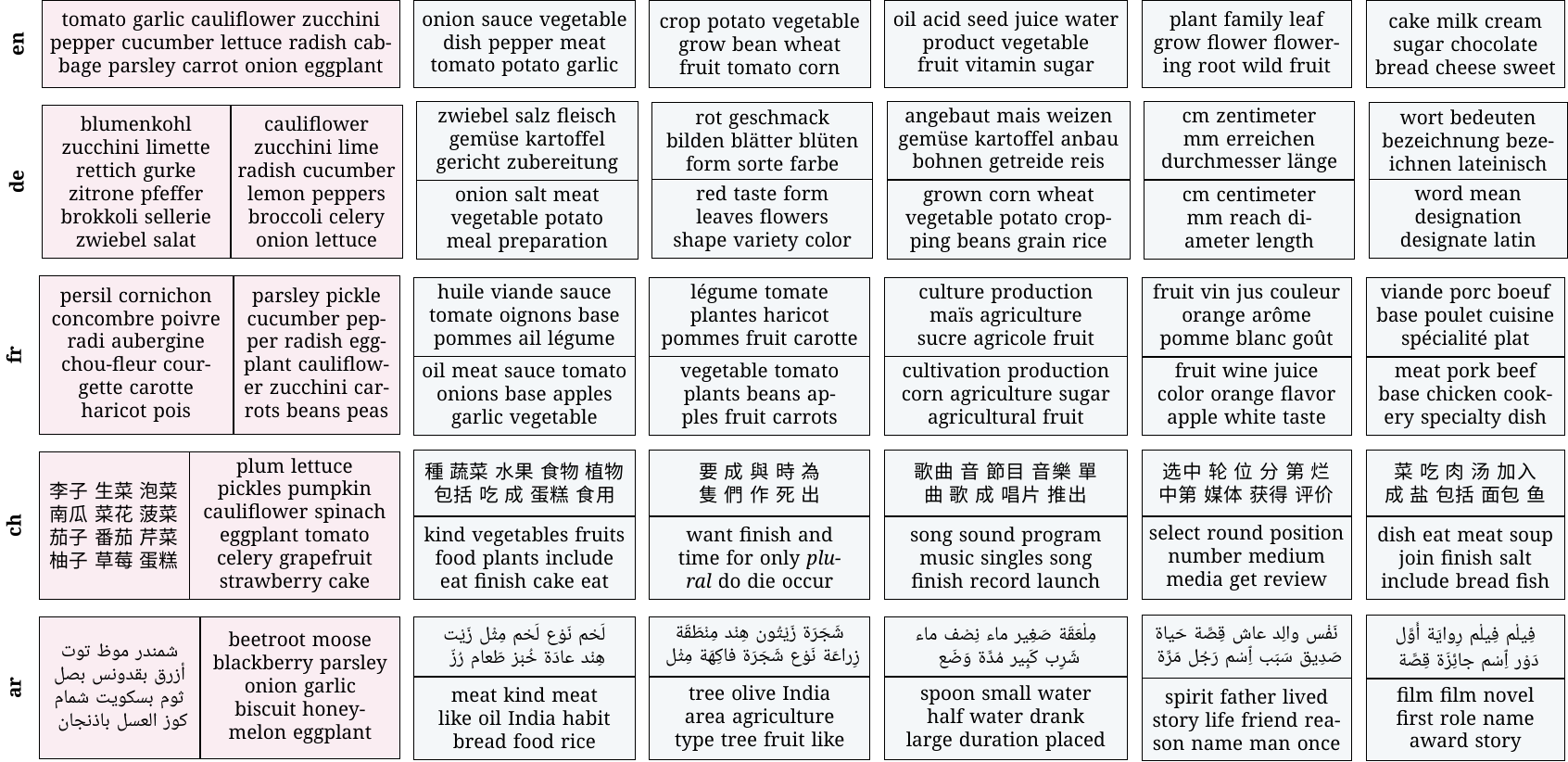}
	\caption{Category \category{vegetables}}
        \label{fig:multilingual_veg}
    \end{subfigure}
    \caption{Categories \category{clothing} (a) and
      \category{vegetables} (b) (light red), and their five most highly
      associated feature types (light blue) for English (en), German
      (de), French (fr), Arabic (ar), and Chinese (ch).  Model output
      of languages other than English was translated into English by
      native speakers.}
\label{fig:multilingual}
\end{figure*}

We observe a slight drop in performance for languages other than
English which is likely due to smaller stimuli sets (see
Table~\ref{tab:corpus:sizes}). BCF, nevertheless, achieves purity
scores of 0.4 or higher for all languages, meaning that on average at
least 40\% of the members of a gold standard category are clustered
together by BCF (purity rises to~58\% for English). This indicates
that meaningful categories emerge throughout.  Qualitative model
output shown in Figures~\ref{fig:bcfenglish} (English)
and~\ref{fig:multilingual} (all languages) corroborates this
result. The categories shown are intuitively meaningful; in particular
\category{vegetable} and \category{clothing}
(Figure~\ref{fig:multilingual}) are interpretable, and thematically
consistent across languages.

A few interesting idiosyncrasies emerge from our cross-lingual
experimental setup, and the ambiguities inherent in language. For
example, the English concepts \concept{tongue} and \concept{bookcase}
were translated into French words \concept{langue} and
\concept{biblioth\`{e}que}, respectively. The French BCF model induced
a category consisting of only these two concepts with highly
associated feature types \{story, author, publish, work, novel\} and
\{meaning, language, Latin, German, form\}. Although this category
does not exist in the gold standard, it is arguably a plausible
inference.  Another example concerns the concept \concept{barrel},
which in the English BCF output, is grouped together with concepts
\concept{cannon, bayonet, bomb} and features like \{kill, fire,
attack\}. In French, on the other hand, it is grouped with
\concept{stove, oven} and the features \{oil, production, ton, gas\}.

We showed that BCF learns meaningful categories across languages which
are quantitatively better than those inferred by a simpler
co-occurrence model. Although generally consistent, categories are
sometimes influenced by characteristics of the respective training and
test language. While the literature confirms an influence of language
on categorization~\citep{Gopnik:1987,Waxman:1995}, this effect is
undoubtedly amplified through our experimental framework.


\section{Experiment 2: Feature Quality}
\begin{table}[ht!]
\begin{center}
\begin{tabular}{clcccc}
 \hline
                      & model            & pr@1 & pr@10 & pr@20 & avg rank \\\hline
\multirow{5}{*}{English}   & BCF      & 0.07 & 0.34  & 0.48  & 61.9 \\
                      & BayesCat & 0.05 & 0.31  & 0.44  & 63.6  \\
                      & Strudel  & 0.04 & 0.31  & 0.45  & 76.1 \\\cdashline{2-6}
		      & co-occ baseline & 0.013 & 0.210 & 0.370  & 97.4 \\
		      & random baseline & 0.002 & 0.020 & 0.040& --\\\hline
\multirow{4}{*}{German}   & BCF      & 0.07 & 0.31  & 0.42  & 90.5  \\
                      & BayesCat & 0.07 & 0.32  & 0.41  & 66.3 \\\cdashline{2-6}
		      & co-occ baseline & 0.007 & 0.110 & 0.137  & 177.6 \\
		      & random baseline & 0.002 & 0.021 & 0.041& --\\\hline
\multirow{4}{*}{French}   & BCF      & 0.07 & 0.31  & 0.44  & 73.2 \\
                      & BayesCat & 0.04 & 0.28  & 0.45  & 64.2 \\\cdashline{2-6}
		      & co-occ baseline & 0.017 & 0.153 & 0.227  & 136.6 \\
		      & random baseline & 0.002 & 0.021 & 0.041& --\\\hline
\multirow{4}{*}{Chinese}   & BCF      & 0.09 & 0.37  & 0.49  & 66.0  \\
                      & BayesCat & 0.09 & 0.38  & 0.53  & 41.7 \\\cdashline{2-6}
		      & co-occ baseline & 0.033 & 0.157 & 0.187  & 139.1 \\
		      & random baseline & 0.002 & 0.022 & 0.044& --\\\hline
\multirow{4}{*}{Arabic}   & BCF      & 0.13 & 0.49  & 0.61  & 54.9  \\
                      & BayesCat & 0.14 & 0.54  & 0.65  &27.5   \\\cdashline{2-6}
		      & co-occ baseline & 0.012 & 0.110 & 0.139  & 154.8 \\
		      & random baseline & 0.003 & 0.025 & 0.050& --\\\hline

\end{tabular}
\caption{Model performance on the concept prediction task in terms of
  precision at rank 1, 10, 20, and average rank
  assigned. We compare BCF (this work), BayesCat,
  a co-occurrence baseline, Strudel (for English only), and a random baseline. 
Results are reported for English, German, French, Chinese and Arabic.}
\label{tab:featurequality}
\end{center}
\end{table}
We next investigate the quality of the features our model learns. We
do this by letting the model predict the right concept solely from a
set of features.  If the model has acquired informative features, they
will be predictive of the unknown concept. Specifically, the model is
presented with a set of previously unseen test stimuli with the target
concept removed.  For each stimulus, the model predicts the missing
concept based on the features $\mathbf{f}$ (i.e., context words).


\subsection{Method}
Like in the category evaluation above, we compare the ranking
performance of BCF, BayesCat, the co-occurrence based model, and
Strudel for English. For the Bayesian models, we directly exploit the
learnt distributions. For BCF, we compute the score of a target
concept~$c$ given a set of features as:

\begin{equation}
Score(c|\mathbf{f}) = \sum_g P(g|c) P(\mathbf{f}|g). 
\end{equation}
Similarly, for BayesCat we compute the score of a concept~$c$ given a set of 
features as follows:  

\begin{equation}
Score(c|\mathbf{f}) = \sum_k P(c|k) P(\mathbf{f}|k).
\end{equation}
For both Strudel and the co-occurrence model, we rank concepts
according to the cumulative association over all observed features for
a particular concept~$c$. For Strudel, association corresponds to
log-likelihood ratio-based association scores, while for the
co-occurrence model it corresponds to co-occurrence counts,
concept~$c$:

\begin{equation}
Score(c|\mathbf{f}) = \sum_{f \in \mathbf{f}} association(c,f).
\end{equation}
We also report a baseline which randomly selects target concepts from
the full set of concepts.

We report precision at rank 1, 10, and~20. We also report the average
rank assigned to the correct concept. All results are based on a
random test set of previously unseen stimuli. 

\begin{figure}[ht]
\begin{center}
\begin{tabular}{|llp{1.5cm}p{1.5cm}p{1.8cm}p{1.5cm}p{1.5cm}|}\hline
\multicolumn{1}{|r}{\textbf{\textit{clam}}} && \multicolumn{5}{p{10cm}|}{\feature{saltwater soft naked marine family}} \\\hline
BCF      &\hspace{0.5cm}\textcolor{white}{a}& snail & \textbf{clam} & moth  & hare & salamander  \\
BayesCat & &  snail            & mackerel & \textbf{clam} & squid & crab  \\
Strudel  & & house & apartment & snail & moth & beetle \\
Co-occ baseline & & level & snail & otter & whale & dolphin \\\hline
\multicolumn{7}{c}{}\\\hline
\multicolumn{1}{|r}{\textbf{\textit{nylon}}} && \multicolumn{5}{p{8cm}|}{\feature{product parachute stocking silk fibre}} \\\hline
BCF  & & {\bf nylon} & crayon & mat & birch & cedar \\
BayesCat & & card & myg & bottle & doll & bag \\
Strudel & & {\bf nylon} & apple & spider & carpet & knee\\
Co-occ baseline & & skirt & nightgown & blouse & necklace & swimsuit \\\hline
\multicolumn{7}{c}{}\\\hline
\multicolumn{1}{|r}{\textbf{\textit{lamb}}} && \multicolumn{5}{p{8cm}|}{\feature{faithful consecrate service gift}} \\\hline
BCF  & & candle & bouquet & bread & robe & {\bf lamb} \\
BayesCat & & cabinet & board & bin & level & whip \\
Strudel & & train & bus & church & ship & chapel\\
Co-occ baseline & & emu & ambulance & train & hook & trolley\\\hline
\end{tabular}
\caption{Illustration of the concept prediction task; we show the
  input presented to the model (i.e.,~features) followed by its
  predictions (target concept). The top five predictions for BCF,
  BayesCat, Strudel, and the co-occurrence model are shown (left to
  right) and the correct answer is highlighted in bold face (if
  included in the top five predictions). }
\label{fig:featurequality}
\end{center}
\end{figure}

\subsection{Results}
\label{sim2-results}
Figure~\ref{fig:featurequality} depicts three English stimuli,
together with concept predictions from BCF and the co-occurrence
model. Table~\ref{tab:featurequality} shows quantitative results of
the three models averaged over a corpus of~300 test stimuli for all
languages. Both BCF and the co-occurrence model outperform the random
baseline by a large margin, and BCF achieves consistently highest
scores.  Both Bayesian models (BCF and BayesCat) outperform the
co-occurrence model across all metrics and conditions. We assume that
plain concept-feature co-occurrence information might be too sparse to
provide a strong signal of concept relevance given a set of
features. The Bayesian models, on the other hand, learn complex
correspondences between features and all concepts in a
category. BayesCat and BCF perform comparably given that they exploit
local co-occurrence relations in similar ways.  BCF learns feature
associations which
discriminate concepts more accurately, suggesting that the joint
learning objective and {\it structured} feature information is
beneficial. The example predictions in Figure~\ref{fig:featurequality}
 corroborate this.

Cross-lingual comparisons reveal that, compared to BCF, the performance of the co-occurrence model degrades more severely for languages other than English. This suggests that BCF can leverage information more efficiently from smaller learning corpora (see~Table~\ref{tab:corpus:sizes}). The number of concepts (i.e.,~target items to be ranked) differs across languages so that absolute numbers are not directly comparable. 

Figures~\ref{fig:bcfenglish} and~\ref{fig:multilingual} qualitatively
support the claim that BCF learns meaningful features across
languages, which are overall coherent and relevant to their associated
category. Some interesting cultural differences emerge, for example
German is the only language for which a \feature{measurement} feature
type is induced for \category{vegetables}
(Figure~\ref{fig:multilingual_veg}; de, 4th from left), while for
\category{clothing}, a \feature{fashion industry} feature emerges in
French (Figure~\ref{fig:multilingual_clo}; fr, 3rd from left). For the
same category, a feature type pertaining to \feature{colour} emerges
for all five languages (\ref{fig:multilingual_clo}, bold margins). In
addition, some features in other languages were not straightforwardly
translatable into English. For example, the 3rd feature type for
\category{vegetables} in Chinese (Figure~\ref{fig:multilingual_veg})
includes the word
\begin{CJK}{UTF8}{gbsn}
 分
\end{CJK} which refers to {\it the extent to which food is cooked} \footnote{Roughly equivalent to the English {\it rare, medium, well-done}, but applicable to all kinds of food.}  and
\begin{CJK}{UTF8}{gbsn}
 烂
\end{CJK} which is {\it the stage when food starts to fall apart after
  cooking (stewing)}. In addition, the feature types induced
for the Chinese \category{clothing} category include two words which both
translate to the English word {\it wear}, but in Chinese are specific
to {\it wearing small items} (e.g.,~jewelery;
\begin{CJK}{UTF8}{gbsn}戴\end{CJK}), and {\it wearing clothes}
(\begin{CJK}{UTF8}{gbsn}穿\end{CJK}), respectively. Language-specific
features are meaningful, and at the same time category-feature
associations across languages reflect culture-driven differences.

\section{Experiment 3:  Feature Relevance and Coherence}
\label{exp-3}
\begin{figure}[t]
\centering
\begin{tabular}{|p{0.12\linewidth}p{0.12\linewidth}p{0.12\linewidth}p{0.13\linewidth}p{0.13\linewidth}p{0.13\linewidth}|}
\hline
\multicolumn{6}{|c|}{\bf `Select the intruder word.'}\\\hline\hline
 \centering$\circ$ & \centering$\circ$ & \centering$\circ$ & \centering$\circ$ & \centering$\bullet$ & {\centering$\circ$}\\
  \centering{\tt color}&\centering{\tt green}&\centering{\tt blue}&\centering{\tt white}&\centering{\tt milk}&{\centering{\tt red}}\\\hline
  \hline
  \centering$\circ$ &\centering$\bullet$ & \centering$\circ$ & \centering$\circ$ &\centering $\circ$ & {\centering$\circ$}\\
  \centering{\tt cell}&\centering{\tt violin}&\centering{\tt study}&\centering{\tt protein}&\centering{\tt human}&{\centering{\tt disease}}\\\hline
\end{tabular}
   \caption{Example
          task for our feature coherence study, correct answers are marked with a filled circle. The example task 
          involves categories and features induced by BCF from the
          English Wikipedia.}
\label{fig:mturkresults_coherence_example}
\end{figure}

\begin{figure}[t]
\centering
\vspace{.2cm}
\begin{tabular}{|p{0.3\linewidth}cp{0.6\linewidth}|}
 \hline
 \multicolumn{3}{|c|}{\bf `Select intruder feature type (right) wrt category (left).'}\\\hline\hline
 \multirow{2}{\linewidth}{\textbf{{\textit{wasp ant caterpillar hornet moth housefly beetle honeydew grasshopper}}}} 
 & $\circ$& \tt insect beetle family larva spider\\
  &$\circ$& \tt tree leaf plant nest grow\\
  &$\bullet$& \tt guitar piano clarinet flute trumpet\\
  &$\circ$& \tt male female egg length cm\\
  & $\circ$& \tt white brown dark tail color\\
  &$\circ$& \tt population habitat bird forest water\\\hline
 \end{tabular}
 \caption{An example
          task for our feature relevance study. The example task
          involves categories and features induced by BCF from the
          English Wikipedia.}
\label{fig:mturkresults_relevance_example}
\end{figure}

Given that our aim is to induce {\it cognitive} representations of the
world, the ultimate assessment of the model's representations is their
meaningfulness to humans, i.e.,~speakers of our target languages. To
this end, we elicited judgments of feature quality from native
speakers using the crowd sourcing platforms
CrowdFlower\footnote{\url{https://www.crowdflower.com/}} and Amazon
Mechanical Turk.\footnote{\url{https://www.mturk.com/}} Specifically,
we are interested in two questions: (1) Do induced feature types have
a single coherent underlying theme such as \feature{color} or
\feature{function} ({\it feature coherence}); (2) Do feature types
associated with a category relate to that category ({\it feature
  relevance})?

We compared the feature types learnt by BCF against the co-occurrence
model as well as BayesCat. For English we also include Strudel. We
omitted the random baseline from this evaluation since it was clearly
inferior in previous simulations.

\subsection{Method}
We adopted the topic intrusion experimental
paradigm~\citep{Chang:2009} for assessing the induced features in two
ways. Firstly, we examined whether the feature types our model learns
are thematically {\it coherent}. Participants were presented features
types (as lists of words), which were augmented with a random
`intruder' feature, and their task was to correctly identify the intruder feature. Figure~\ref{fig:mturkresults_coherence_example} displays an example task. If
the feature types are internally coherent we expect annotators to
identify the intruder with high accuracy. We evaluated all 50 feature
types as induced by BCF and the co-occurrence model. 

Secondly, we assessed the {\it relevance} of feature types assigned
to any category. An example task is shown in~Figure~\ref{fig:mturkresults_relevance_example}. We presented participants with a category and five
feature types (each as a list of words), one of which was randomly
added and was not associated with the category in the model output. Again, they needed to select the correct intruder. If category-feature
type associations induced by the model are generally relevant,
annotators will be able to identify the intruder with high accuracy.
We evaluated all 40 induced categories and their associated features for BCF and
the co-occurrence model. 

For both elicitation studies, we obtained 10 responses per task (see
Figures~\ref{fig:mturkresults_coherence_example}
and~\ref{fig:mturkresults_relevance_example}); participants judged a
single concept and its features per task. All
participants were required to be native speakers of the language they
were evaluating, and we filtered crowdworkers through their location
of residence and self-reported native language (using the
functionality provided by the crowdsourcing platforms). We
additionally included test questions among tasks for which the true
answer was known, and discarded the data from participants who failed
to achieve high accuracy on these test questions. Overall, we
  obtained 50$\times$10 responses for the feature coherence study and
  40$\times$10 responses for feature relevance.

We report the average accuracy across participants of selecting the correct
intruder feature and intruder feature type, respectively. In addition we report 
inter annotator agreement (IAA) using Fleiss Kappa~\citep{Fleiss:1981}. The 
extent to which annotators agree in their judgments allows us to evaluate the 
difficulty of the task, as well as the reliability of the results.

\subsection{Results}
\begin{table}[ht!]
\centering
\begin{tabular}{llcc}
\hline
 & model & accuracy & IAA\\\hline
\multirow{4}{*}{English}& BCF & 0.752 & 0.700\\
& BayesCat & 0.605& 0.456\\
& strudel & 0.435 & 0.364\\\cdashline{2-4}
& co-occ baseline & 0.302 & 0.229\\\hline
\multirow{3}{*}{German}&BCF & 0.530 & 0.361\\
& BayesCat & 0.370& 0.206\\\cdashline{2-4}
& co-occ baseline & 0.340 & 0.201\\\hline
\multirow{3}{*}{French}&BCF & 0.555 & 0.427\\
& BayesCat & 0.468& 0.294 \\\cdashline{2-4}
& co-occ baseline & 0.278 & 0.216\\\hline
\multirow{3}{*}{Chinese}&BCF & 0.468 & 0.419\\
& BayesCat & 0.261 &0.108 \\\cdashline{2-4}
& co-occ baseline & 0.390 & 0.310\\\hline
\multirow{3}{*}{Arabic}&BCF & 0.385 & 0.278\\
& BayesCat & 0.385& 0.215\\\cdashline{2-4}
& co-occ baseline & 0.278 & 0.155\\\hline
\end{tabular}        \caption{Results of our feature relevance study for BCF 
(this work), BayesCat, the co-occurrence
          model (co-occ), and Strudel (for English only) in terms of 
accuracy (the proportion of
          intruders identified correctly) and inter-annotator
          agreement (IAA; Fleiss Kappa~\citep{Fleiss:1981}) for five
          target languages.}
\label{tab:mturkresults_relevance}
\end{table}
Table~\ref{tab:mturkresults_relevance} displays the results for the
{\it feature relevance} study and
Table~\ref{tab:mturkresults_coherence} the {\it feature coherence}
study.  Table~\ref{tab:mturkresults_relevance} shows that humans are
able to detect intruder feature types with higher accuracy in the
context of BCF-induced representations, compared with all comparison
models. Additionally, inter annotator agreement (IAA) is consistently
higher for BCF, indicating that participants more frequently agreed on
their selections and that selecting intruders in the BCF output was an
easier task for them compared to the comparison models. Similar to the
previous simulations, we observe that both Bayesian models (BayesCat
and BCF) outperform the count-based models. In this evaluation,
however, we also observe a clear advantage of BCF compared to
BayesCat, which does not learn structured feature types
inherently. BCF learns to associate {\it relevant} features to
categories.
\begin{table}[ht!]
\centering
\begin{tabular}{llcc}
\hline
& model & accuracy & IAA\\\hline
\multirow{4}{*}{English}& BCF & 0.814 & 0.710\\
& BayesCat & 0.642 & 0.529\\
& strudel & 0.260 & 0.294\\\cdashline{2-4}
& co-occ baseline & 0.266 & 0.296\\\hline
\multirow{3}{*}{German}& BCF & 0.760 & 0.639\\
& BayesCat & 0.465 & 0.341\\\cdashline{2-4}
& co-occ baseline & 0.220 & 0.210\\\hline
\multirow{3}{*}{French}& BCF & 0.690 & 0.527\\
& BayesCat & 0.557 & 0.409\\\cdashline{2-4}
& co-occ baseline & 0.224 & 0.227\\\hline
\multirow{3}{*}{Chinese}& BCF & 0.574 & 0.723\\
& BayesCat & 0.269 & 0.234\\\cdashline{2-4}
& co-occ baseline & 0.228 & 0.227\\\hline
\multirow{3}{*}{Arabic}& BCF & 0.594 & 0.444\\
& BayesCat & 0.416 & 0.423\\\cdashline{2-4}
& co-occ baseline & 0.236 & 0.195\\\hline

\end{tabular}        \caption{Results from our feature coherence study for BCF 
(this work) , BayesCat, the co-occurrence
          model (co-occ), and Strudel (for English only) in terms of 
accuracy (the proportion of intruders identified
          correctly) and inter-annotator agreement (IAA; Fleiss
          Kappa~\citep{Fleiss:1981}) for five target languages.}
\label{tab:mturkresults_coherence}
\end{table}

Table~\ref{tab:mturkresults_coherence} shows the results of the
feature coherence study, where the overall pattern of results is
similar as above. We can see that participants are able to detect
intruder features from the types learnt by BCF more reliably than from
those learnt by all comparison models. Again, both Bayesian models
outperform the count-based baselines both in terms of accuracy and
inter annotator agreement. The superior performance of BCF compared to
BayesCat indicates that its ability to learn structured features
jointly with categories in a single process leads to higher quality
feature representations. In particular, in addition to associating
{\it relevant} feature types with categories, the feature types
themselves are internally coherent, pertaining to different aspects or
properties of the reference category.

Comparing results across languages we observe that scores for English
exceed scores for all other languages. At the same time, for almost
all models and languages the IAA scores fall under the category of
`fair agreement' ($0.20 < \kappa < 0.40$) indicating that the
elicitation task was feasible for crowdworkers. This applies to both
evaluations (Tables~\ref{tab:mturkresults_relevance}
and~\ref{tab:mturkresults_coherence}). We observed a similar pattern
in the results of Experiment~1 (Table~\ref{tab:categoryquality}). We
believe there are two reasons for this drop. Firstly, in order to
perform cross-linguistic experiments, we translated English categories
into other languages. As mentioned in Sections~\ref{sim1-results}
and~\ref{sim2-results}, such a direct correspondence may not always
exist. Consequently, annotators for languages other than English are
faced with a noisier (and potentially harder) task. Secondly, while it
is straightforward to recruit English native speakers on crowd
sourcing platforms, it has proven more challenging for the other
languages. We suspect that our effort to recruit native speakers,
might not have been entirely fail-safe for languages other that
English, and that the language competence of those crowdworkers might
have impacted the quality of their judgments.

Overall, we conclude that {\it jointly} inducing structured features
together with categories from natural language corpora in different
languages enables BCF to learn feature types which are (1)~internally
coherent, referring to a single underlying theme; and (2)~informative
about the categories with which they are associated.

\textbf{}

\section{General Discussion}
\label{sec:general-discussion}

We presented the first large-scale, cross-linguistic analysis of
categorization using naturally occurring data. We showed that rational
Bayesian models of categorization can learn meaningful categories and
their features from complex environments resembling the natural world
more closely than limited laboratory settings.


We developed BCF, a cognitively motivated Bayesian model, and
investigated its ability to learn categories (for hundreds of
concepts) and their structured features from corpora in five
languages.  Like humans `in the wild', our model learns categories and
relevant features {\it jointly}~\cite{Goldstone:2001,Schyns:1997}, and
induces {\it structured } representations of
categories~\cite{Ahn:1998,McRae:2005,Spalding:2000}. Compared to a
simpler co-occurrence model and a Bayesian model with no access to these 
mechanisms BCF learns better categories and features which are rated as 
more relevant and coherent by humans. BCF models category acquisition as a
general, language-independent process. It neither utilizes
language-specific knowledge nor requires language-specific tuning, and
as such paves the way for future investigations involving more
languages, or different kinds of corpora. 

Our study sheds light on the acquisition of concrete concepts and their 
features from text, and as such adopts a constrained view of both the learning 
environment and the learning target. It suggests a number of interesting 
suggestions for future research. First, this article considered natural 
language input as an approximation of the environment from which categories and 
their  representations are learnt. While we showed that the linguistic 
environment is a useful approximation of the full multimodal input a learner has 
access to, it is clear that language cannot capture this multimodal environment 
is not captured in its entirety. Computational models of word learning have been 
trained on multimodal input data (albeit on smaller-scale problems; 
\citep{Frank:2009, Yu:2007}). Advantageously, Bayesian models are flexible with 
respect to the input data they receive, so we expect the application of our 
model to multimodal data to be a feasible avenue for future work. Applying our 
models to such data sets would allow to compare the category acquisition process 
as well as the acquired representations from multimodal input against those 
emerging from language data alone.

A second direction for future work concerns the cognitive assumptions 
underlying the learning setup. The models discussed in this article learn from 
collections of natural language stimuli consisting of a target concept mention 
and its surrounding context. This input is based on the rather bold assumption 
that the learner already has substantial linguistic prior knowledge prior 
to concept and feature learning: she has successfully mapped each target 
concept to a word. As supported by an extensive literature 
\citep{Gopnik:1987, Waxman:1995, borovsky:learning}, word learning, itself a 
fundamental challenge for young infants, and concept learning exhibit a mutual 
influence. Our work remains agnostic about the fact that the meaning of words 
itself needs to be acquired, and that knowledge about concepts and categories 
will help tackle the word learning problem. A fully faithful model would 
consider the problems of word and concept or category learning jointly. 
Extending BCF to account for this joint optimization, and investigating 
emerging acquisition patterns across different languages, will be a very 
interesting avenue for future research. 

Humans not only categorize the physical world around them, but also infer 
complex representations of abstract categories and concepts such as 
\category{political} (e.g., \concept{parliament}, \concept{socialist}), 
\category{legal} (e.g., \concept{law}, \concept{trial}), or \category{feelings} 
(e.g., \concept{mirth} or \concept{embarrassment}). Lacking any physical 
realization, and hence perceivable properties, there is evidence that language 
plays a particularly important role in acquiring the meaning of such abstract 
concepts \citep{Wiemer:2000}. A data-driven study across languages would be 
particularly interesting in the context of abstract categories, whose 
representations are expected to be more sensitive to the cultural environment.

In conclusion, our investigations in to category and feature learning from text 
across languages corroborate prior results~\cite{Riordan:2011} that the 
non-linguistic learning environment is to some extent encoded in language. They 
additionally provide evidence for the stronger statement that the {\it 
structure} of the world which affords rich mental categorical representations is 
encoded in language. We envision scalable testbeds which combine naturally 
occurring data from multiple modalities, for example combining text data with 
images or video. Our work exemplifies the potential of interpretable 
statistical models for gaining insights into the mechanisms which are at play 
in human cognition. We demonstrated the potential of large naturalistic 
datasets for the development and testing of computational models, and are 
confident that computational cognitive models together with large naturally 
occurring data set will open up novel opportunities for investigating human 
cognition at scale.

\section*{Acknowledgments}
This research was funded by the European Research Council (award
number 681760). The funding body had no involvement in the study
design, data collection, analysis and interpretation. It was also not
involved in the writing of the report and the decision to submit the
article for publication.

\bibliographystyle{model1-num-names}
\bibliography{cognition_latest.bib}







\end{document}